\pdfoutput=1
\documentclass[11pt]{article}

\usepackage[final]{acl}

\usepackage{times}
\usepackage{latexsym}

\usepackage[T1]{fontenc}
\usepackage[utf8]{inputenc}

\usepackage{microtype}

\usepackage{inconsolata}

\usepackage{graphicx}

\usepackage{mdframed}
\usepackage{listings}

\usepackage{framed}
\usepackage{verbatim}
\usepackage{fvextra}

\usepackage{booktabs}

\title{Enhancing Function-Calling Capabilities in LLMs: Strategies for \\Prompt Formats, Data Integration, and Multilingual Translation}

\author{%
\textbf{Yi-Chang Chen} \quad \textbf{Po-Chun Hsu} \quad \textbf{Chan-Jan Hsu} \quad \textbf{Da-shan Shiu} \\
\\
MediaTek Research \\
\texttt{\{yi-chang.chen, pochun.hsu, chan.hsu, ds.shiu\}@mtkresearch.com}
}

\begin{document}
\maketitle
\begin{abstract}
Large language models (LLMs) have significantly advanced autonomous agents, particularly in zero-shot tool usage, also known as function calling. This research delves into enhancing the function-calling capabilities of LLMs by exploring different approaches, including prompt formats for integrating function descriptions, blending function-calling and instruction-following data, introducing a novel Decision Token for conditional prompts, leveraging chain-of-thought reasoning, and overcoming multilingual challenges with a translation pipeline. Our key findings and contributions are as follows: (1) Instruction-following data improves both function-calling accuracy and relevance detection. (2) The use of the newly proposed Decision Token, combined with synthetic non-function-call data, enhances relevance detection. (3) A tailored translation pipeline effectively overcomes multilingual limitations, demonstrating significant improvements in Traditional Chinese. These insights highlight the potential for improved function-calling capabilities and multilingual applications in LLMs.
\end{abstract}

\section{Introduction}

The field of autonomous agents has seen remarkable advancements in recent years, largely driven by the capabilities of large language models (LLMs). These models have significantly enhanced the performance of autonomous agents across a variety of tasks \cite{huang-etal-2024-planning-creation,qin2024toolllm,Changle2024tool}. A critical ability for these agents is zero-shot tool usage, also known as function calling. This capability allows LLMs to access up-to-date information from the internet or in-house databases and leverage third-party services, enabling integration with various systems. Such capabilities open up numerous potential applications, including electronic design automation \cite{zhong2023llm4edaemergingprogresslarge}, financial reporting \cite{theuma2024equippinglanguagemodelstool}, and travel planning \cite{Yilun2024travel}.

Despite the progress made through tuning-based methods \cite{grattafiori2024llama3herdmodels,liu2024toolacewinningpointsllm,liu2024apigen} for enabling function-calling capabilities, there remains a gap in research regarding the format variance of prompts, the combination of function-calling data with instruction-following data, and multilingual limitations. This work aims to address these gaps by investigating the following aspects:

\textbf{Prompt Formats:} We explore two strategies for incorporating function descriptions into prompts: (1) introducing a dedicated role for presenting function descriptions, and (2) embedding function descriptions within the system role alongside usage instructions. We aim to determine the impact of these formats on function-calling performance.

\textbf{Data Integration:} We examine the combination of function-calling data with instruction-following data to assess its impact on both instruction-following and function-calling capabilities. Our findings indicate that the use of instruction-following data significantly enhances function-calling accuracy and relevance detection.

\textbf{Decision Token:} We propose a novel Decision Token for conditional prompts, designed to improve relevance detection and facilitate the creation of synthetic non-function-call data for fine-tuning. Our results show that the inclusion of the Decision Token and non-function-call data enhances function-calling relevance detection.

\textbf{Chain-of-Thought (CoT) Reasoning:} We incorporate CoT reasoning through a synthetic data pipeline that constructs reasoning descriptions from sequences of conversations and function calls.

\textbf{Multilingual Translation:} We address the multilingual limitations of current function-calling models by introducing a translation pipeline specifically tailored to overcome the challenges of direct translation methods. Our Traditional Chinese experiments confirm this approach's effectiveness.

In summary, this research provides valuable insights into enhancing LLMs' function-calling capabilities and highlights the potential for multilingual applications. The following sections detail our methodology, experiments, and results, demonstrating the effectiveness of our proposed strategies.

\begin{figure*}[t]
\centering
\includegraphics[width=1.0\textwidth, angle=0]{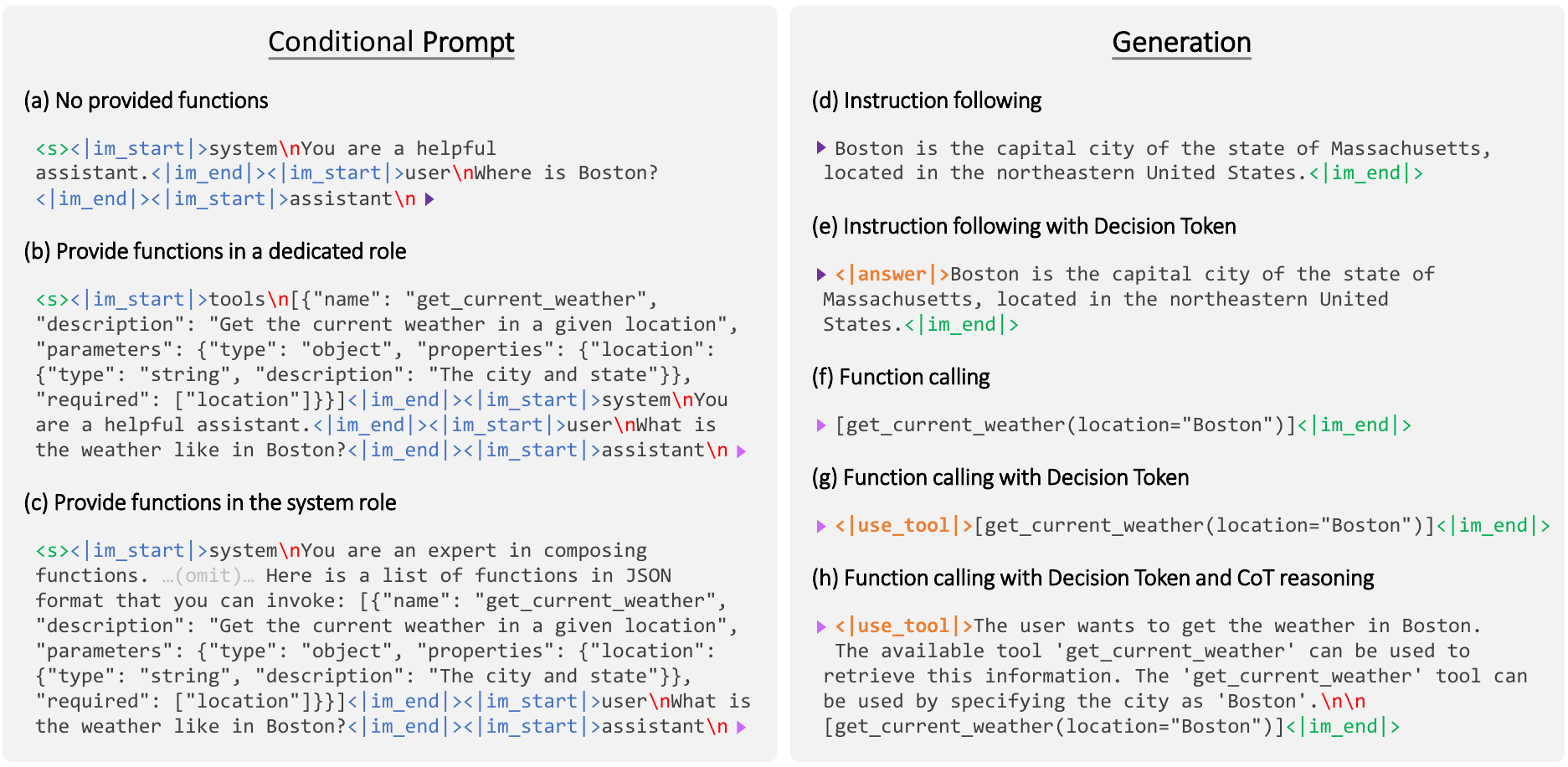}
\caption{
An illustration of prompt templates used for function calling and instruction following in LLMs. During training, LLMs are given conditional prompts (shown on the left) and tasked with generating corresponding text completions (shown on the right). When a function call is required, the model generates structured function calls in the form of a list of functions, where each function is specified with its arguments in the format \texttt{func\_name(arg1=value1, ...)}. Special tokens, including \texttt{<s>}, \texttt{<|im\_start|>}, \texttt{<|im\_end|>}, \texttt{<|answer|>}, and \texttt{<|use\_tool|>}, are each represented by a single token after tokenization. For more details, refer to Section \ref{sec:method-prompt}.
}

\label{fig:prompt-template}
\end{figure*}

\section{Related Work}
Integrating function-calling capabilities into LLMs significantly broadens their problem-solving abilities by enabling interactions with external tools and APIs. Studies have shown that API-integrated LLMs can perform tasks such as programming assistance \cite{gao2022pal}, real-time information retrieval \cite{schick2023toolformer}, complex mathematical computations \cite{heyueya2023solvingmathwordproblems}, and internet utilization \cite{Komeili2021InternetAugmentedDG, gur2024a}. This allows LLMs to access up-to-date information and leverage third-party services, facilitating integration with various systems across advanced applications like electronic design automation \cite{zhong2023llm4edaemergingprogresslarge}, financial reporting \cite{theuma2024equippinglanguagemodelstool}, and travel planning \cite{Yilun2024travel}.

To enable such function-calling capabilities, researchers have explored two main categories of methods. The first involves sophisticated prompting techniques. Frameworks like ReACT \cite{yao2022react} and its successors \cite{xu2023rewoo, shinn2023reflexion, yang2023mmreact, crouse2024formally, wang-etal-2024-llms-imaginarium} combine reasoning and acting within prompts to guide model responses.

More closely related to our work, the second category focuses on training models to generate function calls through fine-tuning. Fine-tuned models such as Gorilla \cite{patil2023gorilla}, ToolAlpaca \cite{tang2023toolalpaca}, ToolLlama \cite{qin2024toolllm}, and the Hermes 3 series by Nous-Research \cite{teknium2024hermes3technicalreport} enhance function-calling capabilities by relying on synthetic data generated by proprietary models like GPT-4 or ChatGPT. Open-source initiatives like NexusRaven-V2 \cite{nexusraven} and IBM's Granite-20B-FunctionCalling \cite{abdelaziz2024granite} aim to develop function-calling models suitable for commercial use without relying on proprietary data. Moreover, many works involve self-supervision to further enhance performance across diverse domains \cite{schick2023toolformer, parisi2022talm, yang2023gpttools, liu2024toolacewinningpointsllm}.

Among the works in the second category, some fine-tuned models and datasets have been openly released. For instance, ToolAlpaca \cite{tang2023toolalpaca} and ToolLLM \cite{qin2024toolllm} have made available their synthetic data or data generation pipelines. ToolACE \cite{liu2024toolacewinningpointsllm} has released both the fine-tuned Llama model and the self-instruction dataset. Additionally, the Gorilla team developed a comprehensive benchmark to evaluate LLMs' function-calling capabilities \cite{berkeley-function-calling-leaderboard}.

Notably, ToolACE \cite{liu2024toolacewinningpointsllm} demonstrated that diversified function-calling sample data helps models learn better function-calling abilities. However, there is a lack of comprehensive analysis on how variations in prompt and meta-information design, as well as the impact of non-function-calling-related instruction tuning data, affect the effectiveness of function-calling capabilities. Existing studies tend to adopt specific prompt templates without extensively investigating the impact of different designs, indicating a need for further research in this area.

% Prompt Formats
% Benchmark (EN and TC)
% Mistral and Breeze

\section{Methodology}

\subsection{Prompt Templates for Function Calling and Instruction Following}
\label{sec:method-prompt}

% Prompt design for Giving Available Functions
% reference to Figure 1

We employ a tuning-based approach to enable both function-calling and instruction-following capabilities in our LLMs. This involves fine-tuning pretrained base models using prompt templates based on the Chat Markup Language (ChatML), a widely adopted format introduced by OpenAI. 

Two main strategies for incorporating function descriptions into prompts are explored: (1) introducing a dedicated role, such as \texttt{tools}, to represent function descriptions in JSON format (Figure \ref{fig:prompt-template}(b)); and (2) embedding function descriptions alongside usage instructions within the system role (Figure \ref{fig:prompt-template}(c)). In the latter strategy, both instruction-following and function-calling are guided by the system prompt. 

During training, the LLMs are provided with conditional prompts as described above and are tasked with generating appropriate text completions. Based on the context, the fine-tuned model dynamically decides whether to respond directly or invoke functions. If no relevant functions are available, the model directly answers the query (Figure \ref{fig:prompt-template}(d)). Otherwise, if function calls are needed, the model generates structured function calls in the form of a list of functions (Figure \ref{fig:prompt-template}(f)).

In the experiments, we investigated the performance comparison of different conditional prompts and the use of training data across various metrics for instruction-following and function-calling capabilities, as discussed in Section \ref{sec:exp-prompt-and-data}.

\subsection{Decision Token}
\label{sec:method-decision-token}

% Next token prediction is a classification task
% Decision before details
% reference to Figure 1

Achieving high performance in relevance detection is challenging, often hindered by the scarcity of negative samples in most synthetic datasets \cite{liu2024toolacewinningpointsllm,liu2024apigen}.

To address this, we propose the novel Decision Token mechanism. LLMs generate responses through next-token prediction, where each step involves a classification task to select the next token. The Decision Token concept leverages the fact that each token prediction is essentially a classification. By introducing a pair of special tokens, the model can predict a binary classification that determines whether to answer the query directly or invoke function calls before generating a detailed response or function calls, respectively. Specifically, this process introduces a pair of special tokens, \texttt{<|answer|>} and \texttt{<|use\_tool|>}, as shown in Figure \ref{fig:prompt-template}(e) and (g). If the model chooses to provide a direct answer, it outputs \texttt{<|answer|>} first; if it chooses function calling, it outputs \texttt{<|use\_tool|>} first. This classification task forces the model to make a decision based on the user query and provided functions before delving into the details, thereby enhancing the stability of its output.

The Decision Token also facilitates the creation of non-function-call data from function-called data. To generate non-function-call data, consider an example where the original data  involves three functions: \texttt{func\_A}, \texttt{func\_B}, and \texttt{func\_C}. Based on the user query, \texttt{func\_A} is helpful and thus called in the original data point. By assuming that \texttt{func\_B} and \texttt{func\_C} are not helpful, we can create non-function-call data by removing \texttt{func\_A} as input. With only \texttt{func\_B} and \texttt{func\_C} as the remaining functions, function calling should not be triggered from user query and a direct answer should be provided. This allows us to easily obtain non-function-call data. Previously, generating non-function-call data for training was challenging because it required specific LLM responses for non-function-call cases. However, with the Decision Token, we can train the model to output only \texttt{<|answer|>} in non-function-call cases. During inference, this is not an issue because the model will continue to provide an appropriate response after \texttt{<|answer|>}.

The experiments involving the Decision Token and training on synthetic non-function-call data are discussed in Section \ref{sec:exp-decision-token}.

% When we adopt the Decision Token, we can also easily create many irrelevant data examples, where function calling is not involved, from function calling involved data. The advantage is that when the state belongs to an irrelevant case, we can design training data to predict only \texttt{<|answer|>} without needing to specify how to answer. During inference, the model can use its own capabilities to provide a proper response after \texttt{<|answer|>}. Specifically, we can generate irrelevant data by identifying the functions involved in function calling data, removing these functions, and assuming that the remaining functions do not help the user query. We then create an irrelevant case by combining the remaining functions with the user query and train the model to predict only the \texttt{<|answer|>} token. These training data will enhance the model's ability in relevance detection. The experiments on Decision Token and training on irrelevant cases are presented in Section \ref{sec:exp-decision-token}.

\subsection{Chain-of-thought Reasoning}
\label{sec:method-cot}

% CoT
% synthetic data pipeline for FC CoT
% reference to Figure 1

CoT reasoning has been demonstrated to significantly enhance performance across various tasks by incorporating intermediate reasoning steps \cite{wei2022chain}. Inspired by this, we explore whether CoT reasoning can similarly improve function-calling capabilities. To achieve this, we propose a synthetic data generation pipeline that constructs reasoning descriptions derived from sequences of conversations and function calls. This pipeline leverages single-turn queries with commercial-grade LLMs. In our prompt design, we initially provide the history of the conversation and the available functions, requiring the identification of the reasoning needed to determine how to use the available functions to achieve the target function calls. Additionally, we provide multiple examples to enhance stability (few-shot learning). More details are provided in Appendix \ref{sec:appendix-reason}. Using this pipeline, we generate data that captures the thinking process, which is then used to fine-tune base LLMs. The fine-tuning process employs a structured prompt template, as illustrated in Figure \ref{fig:prompt-template}(h). The experiments on incorporating CoT reasoning are presented in Section \ref{sec:exp-cot}.

\subsection{Multilingual Translation}
\label{sec:method-translation}

% Directly translation is not correct
% Novel pipeline

To enhance the multilingual capabilities of function-calling tuning, translating existing English function-calling datasets into target languages is a common approach. However, this process presents significant challenges, as elements such as function names, enumeration items, and structured function calls cannot be directly translated without risking inconsistencies or errors. To address these issues and maintain the semantic and syntactic integrity of translated datasets, we propose a novel translation pipeline specifically designed to overcome the limitations of direct translation methods. This pipeline leverages a single-turn query with commercial-grade LLMs. In our prompt design, we initially specify the JSON format of the provided conversation trajectory with function calls. We then instruct the LLMs to translate the data into the target language, adhering to the rules of not translating function names and descriptions, and translating arguments only when reasonable. More details are provided in Appendix \ref{sec:appendix-translate}. The experiments on verifying the effectiveness of the pipeline is presented in Section \ref{sec:exp-translation}.

\begin{table*}[t]
\centering
\begin{tabular}{llcc|ccc}
\toprule
~ & ~ & \multicolumn{2}{c|}{\textbf{Use of Data?}} & \textbf{MT} & \textbf{AST} & \textbf{Relevance}  \\
~ & \textbf{Conditional Prompt} &  \multicolumn{1}{c}{\texttt{IF-110k}} & \multicolumn{1}{c|}{\texttt{FC-110k}} & \textbf{Bench} & \textbf{Summary} & \textbf{Detection}   \\
\midrule
(a) & No provided functions & $\bigcirc$ & $\times$ & 5.46 & - & -  \\
(b) & Provide functions in a dedicated role & $\bigcirc$ & $\bigcirc$ & \textbf{5.57} & 85.25 & \textbf{49.58}  \\
(c) & Provide functions in the system role & $\bigcirc$ & $\bigcirc$ & 5.29 & \textbf{85.94} & 39.58  \\
(d) & Provide functions in a dedicated role & $\times$ & $\bigcirc$ & - & 74.62 & 38.33   \\
(e) & Provide functions in the system role & $\times$ & $\bigcirc$ & - & 74.50 & 27.08   \\
\bottomrule
\end{tabular}
\caption{
Performance comparison of different prompts and the use of data on various metrics for instruction-following and function-calling capabilities. The "Use of Data?" columns indicate whether the respective datasets (\texttt{IF-110k} and \texttt{FC-110k}) are included in the training process. Detailed experiments are discussed in Section \ref{sec:exp-prompt-and-data}.
}
\label{tab:prompt-template}
\end{table*}

\begin{table*}[t]
\centering
\begin{tabular}{l|cc|cc}
\toprule
 \multicolumn{1}{r|}{\textbf{How to provide functions in a prompt?}} & \multicolumn{2}{c|}{\textbf{In a dedicated role}} & \multicolumn{2}{c}{\textbf{In the system role}} \\ 
\midrule
\multicolumn{1}{r|}{\textbf{Metrics on BFCL} \cite{berkeley-function-calling-leaderboard}\textbf{:}} & \textbf{AST} & \textbf{Relevance} & \textbf{AST} & \textbf{Relevance} \\ 
 & \textbf{Summary} & \textbf{Detection} & \textbf{Summary} & \textbf{Detection} \\ 
\midrule
Baseline & \textbf{85.25} & 49.58 & \textbf{85.94} & 39.58 \\ 
+ Decision Token & \textbf{85.25} & 37.50 & 84.63 & 47.50 \\ 
+ Non-function-call Data (\texttt{NF-1k}) & 84.81 & \textbf{57.50} & 83.44 & \textbf{65.42} \\ 
\bottomrule
\end{tabular}
\caption{
Impact of incrementally adding the Decision Token and synthetic non-function-call data. The table shows different prompt configurations for providing functions. The last three rows represent the configurations: baseline, Decision Token added, and both Decision Token and synthetic data added. See Section \ref{sec:exp-decision-token} for details.
}
\label{tab:decision}
\end{table*}

\section{Experiments and Results}

\subsection{Experimental Setup}
\label{sec:setup}

In this section, we describe the experimental setup used to evaluate our proposed methods, including details on datasets, model configurations, training parameters, and evaluation metrics. 

We created a diverse dataset for fine-tuning, which includes both instruction-following and function-calling examples. The instruction-following data, marked as \texttt{IF-110k}, consists of 110k instances sampled from Open ORCA \cite{longpre2023flan}, a synthetic dataset generated from GPT-4 completions. The function-calling data, marked as \texttt{FC-110k}, also includes 110k instances, sourced from a combination of APIGen \cite{liu2024apigen} and the glaive-function-calling-v2 dataset\footnote{https://huggingface.co/datasets/glaiveai/glaive-function-calling-v2}.

We used Breeze-7B\footnote{https://huggingface.co/MediaTek-Research/Breeze-7B-Base-v1\_0} as the base model for our experiments. Breeze-7B \cite{hsu2024breeze} is an open-source language model based on Mistral-7B, designed to improve language comprehension and chatbot capabilities in Traditional Chinese. Using Breeze-7B, we can test the model's effectiveness in both English and Traditional Chinese. 

The models were fine-tuned using the prompt templates, described in Section \ref{sec:method-prompt}. For fine-tuning, we applied the low-rank adaptation (LoRA) technique on linear layers. The fine-tuning process used the following hyperparameters: a learning rate of 1e-4, a batch size of 48, 3 epochs, a cosine learning rate scheduler, the AdamW optimizer, 100 warmup steps, a LoRA rank ($r$) of 16, and a LoRA $\alpha$ of 32.

We evaluated the performance of our models using the following metrics:

\textbf{AST Summary (\%)}: This metric, used in the Berkeley Function Calling Leaderboard (BFCL) \cite{berkeley-function-calling-leaderboard}, assesses the structural correctness of language model outputs for function-calling tasks by comparing the Abstract Syntax Tree (AST) representations of generated and target function calls. It includes four problem types—Simple Function, Multiple Function, Parallel Function, and Parallel Multiple Function—categorized based on the combination of the number of provided functions and function calls. The dataset consists of 400 Simple Function tasks and 200 tasks for each of the other three types. The AST Summary is the average accuracy across these four types.

\textbf{Relevance Detection (\%)}: This metric, also used in the BFCL, measures the success rate of no function call when none of the provided functions are relevant. This scenario helps determine whether a model will hallucinate its functions and parameters when the provided functions are irrelevant to the user's query.

\textbf{MT-Bench (score)}: Unlike previous works, we 
 also explore the impact of instruction-following capabilities when enabling function-calling functionalities. MT-Bench \cite{zheng2023judging} is a benchmark for evaluating these capabilities. We use GPT-4o as a judge to give the score out of 10.

We also evaluated the performance on Traditional Chinese function calling using the Function Calling Leaderboard for ZHTW \cite{fc-leaderboard-zhtw}, which is constructed by translating the BFCL. Therefore, the calculation of metrics AST Summary and Relevance Detection is similar.

\begin{table*}[t]
\centering
\begin{tabular}{l|cc|cc}
\toprule
 \multicolumn{1}{r|}{\textbf{How to provide functions in a prompt?}} & \multicolumn{2}{c|}{\textbf{In a dedicated role}} & \multicolumn{2}{c}{\textbf{In the system role}} \\ 
\midrule
 \multicolumn{1}{l|}{\textbf{Metrics on Function Calling Leaderboard}}  & \textbf{AST} & \textbf{Relevance} & \textbf{AST} & \textbf{Relevance} \\ 
\multicolumn{1}{r|}{~~\textbf{ for ZHTW} \cite{fc-leaderboard-zhtw}\textbf{:}} & \textbf{Summary} & \textbf{Detection} & \textbf{Summary} & \textbf{Detection} \\ 
\midrule
Baseline & 52.37 &36.67 & 50.81 & \textbf{47.08} \\ 
+ Traditional Chinese Data (\texttt{TC-19k}) & \textbf{61.56} &  \textbf{41.25} & \textbf{58.56} & 45.83 \\ 
\bottomrule
\end{tabular}
\caption{
The impact of adding Traditional Chinese data, generated through a tailored translation pipeline (Section \ref{sec:method-translation}), is analyzed. Notably, the metrics AST Summary and Relevance Detection are evaluated on the benchmark for Tradition Chinese. Detailed experiments are discussed in Section \ref{sec:exp-translation}.
}
\label{tab:tc}
\end{table*}

\subsection{Effects of Prompt Templates and Use of Training Data}
\label{sec:exp-prompt-and-data}

We investigated the performance comparison of different conditional prompts and the use of training data on various metrics for instruction-following and function-calling capabilities, as shown in Table \ref{tab:prompt-template}. Conditional prompts are described in Section \ref{sec:method-prompt}. The use of training data, training setup, and metrics is described in Section \ref{sec:setup}.

Compared to Table \ref{tab:prompt-template}(b) and (c), the functions provided in a dedicated role and the system role exhibit similar capabilities in terms of instruction-following (MT Bench) and function-calling accuracy (AST Summary). But, Relevance Detection is superior when functions are provided in the dedicated role. We hypothesize that providing functions in the dedicated role makes the template with functions significantly different from the template without functions, making it easier for the model to learn when to use function calling or respond directly.

Compared to the results shown in Table \ref{tab:prompt-template}(a), (b), and (c) on the MT Bench, we find that enabling the function-calling capability does not reduce the performance of the instruction-following capability, regardless of the conditional prompt given.

Compared to the results shown in Table \ref{tab:prompt-template}(b), (c), (d), and (e) on the AST Summary and Relevance Detection metrics, we find that the performance of the function-calling capability decreases when we exclude the instruction-following data (\texttt{IF-110k}). This observation is noteworthy. We hypothesize that the increase in function-calling capability is due to the additional instruction-following data, which helps the model better understand the semantic structure of the prompts. Consequently, this improved understanding enhances the model's ability to accurately perform function calling. Moreover, instruction-following data provided more non-function-call examples, further improving Relevance Detection.

In conclusion, our experiments demonstrate that the inclusion of function-calling capabilities does not compromise instruction-following performance. Additionally, the use of instruction-following data significantly enhances function-calling accuracy and relevance detection.

\subsection{Effects of the Decision Token}
\label{sec:exp-decision-token}

To verify the effectiveness of the Decision Token, as described in Section \ref{sec:method-decision-token}, we examined the effects of incrementally adding the Decision Token and the synthetic non-function-call data. 

In the baseline experiment, we used \texttt{IF-110k} and \texttt{FC-110k} as the training data to finetune the base model. Then, we added the Decision Token to the prompt templates and finetuned the base model on the same training data. In the final experiment, we used synthetic methods described in Section \ref{sec:method-decision-token} to generate 1k instances of non-function-call data, marked as \texttt{NF-1k}. The models were then finetuned with a combination of \texttt{NF-1k}, \texttt{IF-110k}, and \texttt{FC-110k}. The results of this investigation are presented in Table \ref{tab:decision}. In conclusion, our analysis shows that the adoption of the Decision Token, along with the accompanying synthetic non-function-call data, can benefit Relevance Detection. However, it also results in a slight decrease in function-calling accuracy (AST Summary).

\subsection{Effects of Chain-of-Thought Reasoning}
\label{sec:exp-cot}

To evaluate CoT reasoning (Section \ref{sec:method-cot}), we generated reasoning descriptions for each function call in \texttt{FC-110k}, creating \texttt{FC-110k-Reason}. Comparing models trained on \texttt{IF-110k} + \texttt{FC-110k-Reason} with those trained on \texttt{IF-110k} + \texttt{FC-110k}, we found no significant improvement in function calling accuracy (AST Summary), which was 84.44\% compared to the baseline of 85.25\%. We hypothesize that BFCL problems may not require reasoning for function calling.

% To verify the effectiveness of the CoT reasoning, as described in Section \ref{sec:method-cot}, we used the synthetic methods to generate reasoning descriptions for each function call in \texttt{FC-110k}. The generated data is referred to as \texttt{FC-110k-Reason}. We compared the performance of models trained on \texttt{IF-110k} combined with \texttt{FC-110k-Reason} to those trained on \texttt{IF-110k} combined with \texttt{FC-110k}, and found that the function calling accuracy (AST Summary) did not show the expected improvement. Compared to the baseline AST Summary of 85.25\%, the model trained on data with the added reasoning process achieved 84.44\%. We hypothesize that the lack of expected improvement is due to the nature of the problems in BFCL, which may not require reasoning for function calling.

\subsection{Effects of Translation Pipeline}
\label{sec:exp-translation}

% reference to Table 3
% balance selection

To evaluate the effectiveness of the translation pipeline described in Section \ref{sec:method-translation}, we generated 18k function-calling instances in Traditional Chinese using synthetic methods from the \texttt{FC-110k} dataset. Additionally, we applied an non-function-call case generation pipeline, as detailed in Section \ref{sec:method-decision-token}, to this dataset, producing 200 instances of non-function-call data in Traditional Chinese. The combined dataset is referred to as \texttt{TC-19k}. 

In our baseline experiment, we used the Decision Token approach along with the \texttt{IF-110k}, \texttt{FC-110k}, and \texttt{NF-1k} datasets as training data. We then incorporated the \texttt{TC-19k} Traditional Chinese data into the training set. The results, presented in Table \ref{tab:tc}, demonstrate that even a small amount of translated data can significantly enhance function-calling performance.

\section{Conclusion}

Our research demonstrates that integrating instruction-following data with function-calling tasks significantly enhances function-calling capabilities. The Decision Token mechanism, combined with synthetic non-function-call data, further improves relevance detection. Additionally, a tailored translation pipeline effectively mitigates multilingual challenges. These findings underscore the potential for improving function-calling capabilities and expanding multilingual proficiency in LLMs, paving the way for more practical real-world applications.

% These findings underscore the potential for improving function-calling capabilities and expanding multilingual proficiency in LLMs, paving the way for more practical real-world applications.

% We explored strategies to enhance the function-calling capabilities of LLMs. Our findings show that integrating instruction-following data improves function-calling capabilities, while the Decision Token and synthetic data further boost detection capabilities. A tailored translation pipeline address multilingual limitations. 

% \section*{Acknowledgments}

\bibliography{anthology}

\begin{thebibliography}{33}
\providecommand{\natexlab}[1]{#1}

\bibitem[{Abdelaziz et~al.(2024)Abdelaziz, Basu, Agarwal, Kumaravel, Stallone, Panda, Rizk, Bhargav, Crouse, Gunasekara, Ikbal, Joshi, Karanam, Kumar, Munawar, Neelam, Raghu, Sharma, Soria, Sreedhar, Venkateswaran, Unuvar, Cox, Roukos, Lastras, and Kapanipathi}]{abdelaziz2024granite}
Ibrahim Abdelaziz, Kinjal Basu, Mayank Agarwal, Sadhana Kumaravel, Matthew Stallone, Rameswar Panda, Yara Rizk, GP~Bhargav, Maxwell Crouse, Chulaka Gunasekara, Shajith Ikbal, Sachin Joshi, Hima Karanam, Vineet Kumar, Asim Munawar, Sumit Neelam, Dinesh Raghu, Udit Sharma, Adriana~Meza Soria, Dheeraj Sreedhar, Praveen Venkateswaran, Merve Unuvar, David Cox, Salim Roukos, Luis Lastras, and Pavan Kapanipathi. 2024.
\newblock \href {https://arxiv.org/abs/2407.00121} {Granite-function calling model: Introducing function calling abilities via multi-task learning of granular tasks}.
\newblock \emph{Preprint}, arXiv:2407.00121.

\bibitem[{Crouse et~al.(2024)Crouse, Abdelaziz, Basu, Dan, Kumaravel, Fokoue, Kapanipathi, and Lastras}]{crouse2024formally}
Maxwell Crouse, Ibrahim Abdelaziz, Kinjal Basu, Soham Dan, Sadhana Kumaravel, Achille Fokoue, Pavan Kapanipathi, and Luis~A. Lastras. 2024.
\newblock \href {https://openreview.net/forum?id=FRxDrdysBt} {Formally specifying the high-level behavior of {LLM}-based agents}.

\bibitem[{Gao et~al.(2022)Gao, Madaan, Zhou, Alon, Liu, Yang, Callan, and Neubig}]{gao2022pal}
Luyu Gao, Aman Madaan, Shuyan Zhou, Uri Alon, Pengfei Liu, Yiming Yang, Jamie Callan, and Graham Neubig. 2022.
\newblock Pal: Program-aided language models.
\newblock \emph{arXiv preprint arXiv:2211.10435}.

\bibitem[{Grattafiori et~al.(2024)Grattafiori, Dubey, Jauhri, Pandey, Kadian, Al-Dahle, Letman, Mathur, Schelten, Vaughan, Yang, Fan, Goyal, Hartshorn, Yang, Mitra, Sravankumar, Korenev, Hinsvark, Rao, Zhang, Rodriguez, Gregerson, Spataru, Roziere, Biron, Tang, Chern, Caucheteux, Nayak, Bi, Marra, McConnell, Keller, Touret, Wu, Wong, Ferrer, Nikolaidis, Allonsius, Song, Pintz, Livshits, Wyatt, Esiobu, Choudhary, Mahajan, Garcia-Olano, Perino, Hupkes, Lakomkin, AlBadawy, Lobanova, Dinan, Smith, Radenovic, Guzmán, Zhang, Synnaeve, Lee, Anderson, Thattai, Nail, Mialon, Pang, Cucurell, Nguyen, Korevaar, Xu, Touvron, Zarov, Ibarra, Kloumann, Misra, Evtimov, Zhang, Copet, Lee, Geffert, Vranes, Park, Mahadeokar, Shah, van~der Linde, Billock, Hong, Lee, Fu, Chi, Huang, Liu, Wang, Yu, Bitton, Spisak, Park, Rocca, Johnstun, Saxe, Jia, Alwala, Prasad, Upasani, Plawiak, Li, Heafield, Stone, El-Arini, Iyer, Malik, Chiu, Bhalla, Lakhotia, Rantala-Yeary, van~der Maaten, Chen, Tan, Jenkins, Martin, Madaan, Malo, Blecher,
  Landzaat, de~Oliveira, Muzzi, Pasupuleti, Singh, Paluri, Kardas, Tsimpoukelli, Oldham, Rita, Pavlova, Kambadur, Lewis, Si, Singh, Hassan, Goyal, Torabi, Bashlykov, Bogoychev, Chatterji, Zhang, Duchenne, Çelebi, Alrassy, Zhang, Li, Vasic, Weng, Bhargava, Dubal, Krishnan, Koura, Xu, He, Dong, Srinivasan, Ganapathy, Calderer, Cabral, Stojnic, Raileanu, Maheswari, Girdhar, Patel, Sauvestre, Polidoro, Sumbaly, Taylor, Silva, Hou, Wang, Hosseini, Chennabasappa, Singh, Bell, Kim, Edunov, Nie, Narang, Raparthy, Shen, Wan, Bhosale, Zhang, Vandenhende, Batra, Whitman, Sootla, Collot, Gururangan, Borodinsky, Herman, Fowler, Sheasha, Georgiou, Scialom, Speckbacher, Mihaylov, Xiao, Karn, Goswami, Gupta, Ramanathan, Kerkez, Gonguet, Do, Vogeti, Albiero, Petrovic, Chu, Xiong, Fu, Meers, Martinet, Wang, Wang, Tan, Xia, Xie, Jia, Wang, Goldschlag, Gaur, Babaei, Wen, Song, Zhang, Li, Mao, Coudert, Yan, Chen, Papakipos, Singh, Srivastava, Jain, Kelsey, Shajnfeld, Gangidi, Victoria, Goldstand, Menon, Sharma, Boesenberg,
  Baevski, Feinstein, Kallet, Sangani, Teo, Yunus, Lupu, Alvarado, Caples, Gu, Ho, Poulton, Ryan, Ramchandani, Dong, Franco, Goyal, Saraf, Chowdhury, Gabriel, Bharambe, Eisenman, Yazdan, James, Maurer, Leonhardi, Huang, Loyd, Paola, Paranjape, Liu, Wu, Ni, Hancock, Wasti, Spence, Stojkovic, Gamido, Montalvo, Parker, Burton, Mejia, Liu, Wang, Kim, Zhou, Hu, Chu, Cai, Tindal, Feichtenhofer, Gao, Civin, Beaty, Kreymer, Li, Adkins, Xu, Testuggine, David, Parikh, Liskovich, Foss, Wang, Le, Holland, Dowling, Jamil, Montgomery, Presani, Hahn, Wood, Le, Brinkman, Arcaute, Dunbar, Smothers, Sun, Kreuk, Tian, Kokkinos, Ozgenel, Caggioni, Kanayet, Seide, Florez, Schwarz, Badeer, Swee, Halpern, Herman, Sizov, Guangyi, Zhang, Lakshminarayanan, Inan, Shojanazeri, Zou, Wang, Zha, Habeeb, Rudolph, Suk, Aspegren, Goldman, Zhan, Damlaj, Molybog, Tufanov, Leontiadis, Veliche, Gat, Weissman, Geboski, Kohli, Lam, Asher, Gaya, Marcus, Tang, Chan, Zhen, Reizenstein, Teboul, Zhong, Jin, Yang, Cummings, Carvill, Shepard, McPhie,
  Torres, Ginsburg, Wang, Wu, U, Saxena, Khandelwal, Zand, Matosich, Veeraraghavan, Michelena, Li, Jagadeesh, Huang, Chawla, Huang, Chen, Garg, A, Silva, Bell, Zhang, Guo, Yu, Moshkovich, Wehrstedt, Khabsa, Avalani, Bhatt, Mankus, Hasson, Lennie, Reso, Groshev, Naumov, Lathi, Keneally, Liu, Seltzer, Valko, Restrepo, Patel, Vyatskov, Samvelyan, Clark, Macey, Wang, Hermoso, Metanat, Rastegari, Bansal, Santhanam, Parks, White, Bawa, Singhal, Egebo, Usunier, Mehta, Laptev, Dong, Cheng, Chernoguz, Hart, Salpekar, Kalinli, Kent, Parekh, Saab, Balaji, Rittner, Bontrager, Roux, Dollar, Zvyagina, Ratanchandani, Yuvraj, Liang, Alao, Rodriguez, Ayub, Murthy, Nayani, Mitra, Parthasarathy, Li, Hogan, Battey, Wang, Howes, Rinott, Mehta, Siby, Bondu, Datta, Chugh, Hunt, Dhillon, Sidorov, Pan, Mahajan, Verma, Yamamoto, Ramaswamy, Lindsay, Lindsay, Feng, Lin, Zha, Patil, Shankar, Zhang, Zhang, Wang, Agarwal, Sajuyigbe, Chintala, Max, Chen, Kehoe, Satterfield, Govindaprasad, Gupta, Deng, Cho, Virk, Subramanian, Choudhury,
  Goldman, Remez, Glaser, Best, Koehler, Robinson, Li, Zhang, Matthews, Chou, Shaked, Vontimitta, Ajayi, Montanez, Mohan, Kumar, Mangla, Ionescu, Poenaru, Mihailescu, Ivanov, Li, Wang, Jiang, Bouaziz, Constable, Tang, Wu, Wang, Wu, Gao, Kleinman, Chen, Hu, Jia, Qi, Li, Zhang, Zhang, Adi, Nam, Yu, Wang, Zhao, Hao, Qian, Li, He, Rait, DeVito, Rosnbrick, Wen, Yang, Zhao, and Ma}]{grattafiori2024llama3herdmodels}
Aaron Grattafiori, Abhimanyu Dubey, Abhinav Jauhri, Abhinav Pandey, Abhishek Kadian, Ahmad Al-Dahle, Aiesha Letman, Akhil Mathur, Alan Schelten, Alex Vaughan, Amy Yang, Angela Fan, Anirudh Goyal, Anthony Hartshorn, Aobo Yang, Archi Mitra, Archie Sravankumar, Artem Korenev, Arthur Hinsvark, Arun Rao, Aston Zhang, Aurelien Rodriguez, Austen Gregerson, Ava Spataru, Baptiste Roziere, Bethany Biron, Binh Tang, Bobbie Chern, Charlotte Caucheteux, Chaya Nayak, Chloe Bi, Chris Marra, Chris McConnell, Christian Keller, Christophe Touret, Chunyang Wu, Corinne Wong, Cristian~Canton Ferrer, Cyrus Nikolaidis, Damien Allonsius, Daniel Song, Danielle Pintz, Danny Livshits, Danny Wyatt, David Esiobu, Dhruv Choudhary, Dhruv Mahajan, Diego Garcia-Olano, Diego Perino, Dieuwke Hupkes, Egor Lakomkin, Ehab AlBadawy, Elina Lobanova, Emily Dinan, Eric~Michael Smith, Filip Radenovic, Francisco Guzmán, Frank Zhang, Gabriel Synnaeve, Gabrielle Lee, Georgia~Lewis Anderson, Govind Thattai, Graeme Nail, Gregoire Mialon, Guan Pang,
  Guillem Cucurell, Hailey Nguyen, Hannah Korevaar, Hu~Xu, Hugo Touvron, Iliyan Zarov, Imanol~Arrieta Ibarra, Isabel Kloumann, Ishan Misra, Ivan Evtimov, Jack Zhang, Jade Copet, Jaewon Lee, Jan Geffert, Jana Vranes, Jason Park, Jay Mahadeokar, Jeet Shah, Jelmer van~der Linde, Jennifer Billock, Jenny Hong, Jenya Lee, Jeremy Fu, Jianfeng Chi, Jianyu Huang, Jiawen Liu, Jie Wang, Jiecao Yu, Joanna Bitton, Joe Spisak, Jongsoo Park, Joseph Rocca, Joshua Johnstun, Joshua Saxe, Junteng Jia, Kalyan~Vasuden Alwala, Karthik Prasad, Kartikeya Upasani, Kate Plawiak, Ke~Li, Kenneth Heafield, Kevin Stone, Khalid El-Arini, Krithika Iyer, Kshitiz Malik, Kuenley Chiu, Kunal Bhalla, Kushal Lakhotia, Lauren Rantala-Yeary, Laurens van~der Maaten, Lawrence Chen, Liang Tan, Liz Jenkins, Louis Martin, Lovish Madaan, Lubo Malo, Lukas Blecher, Lukas Landzaat, Luke de~Oliveira, Madeline Muzzi, Mahesh Pasupuleti, Mannat Singh, Manohar Paluri, Marcin Kardas, Maria Tsimpoukelli, Mathew Oldham, Mathieu Rita, Maya Pavlova, Melanie Kambadur,
  Mike Lewis, Min Si, Mitesh~Kumar Singh, Mona Hassan, Naman Goyal, Narjes Torabi, Nikolay Bashlykov, Nikolay Bogoychev, Niladri Chatterji, Ning Zhang, Olivier Duchenne, Onur Çelebi, Patrick Alrassy, Pengchuan Zhang, Pengwei Li, Petar Vasic, Peter Weng, Prajjwal Bhargava, Pratik Dubal, Praveen Krishnan, Punit~Singh Koura, Puxin Xu, Qing He, Qingxiao Dong, Ragavan Srinivasan, Raj Ganapathy, Ramon Calderer, Ricardo~Silveira Cabral, Robert Stojnic, Roberta Raileanu, Rohan Maheswari, Rohit Girdhar, Rohit Patel, Romain Sauvestre, Ronnie Polidoro, Roshan Sumbaly, Ross Taylor, Ruan Silva, Rui Hou, Rui Wang, Saghar Hosseini, Sahana Chennabasappa, Sanjay Singh, Sean Bell, Seohyun~Sonia Kim, Sergey Edunov, Shaoliang Nie, Sharan Narang, Sharath Raparthy, Sheng Shen, Shengye Wan, Shruti Bhosale, Shun Zhang, Simon Vandenhende, Soumya Batra, Spencer Whitman, Sten Sootla, Stephane Collot, Suchin Gururangan, Sydney Borodinsky, Tamar Herman, Tara Fowler, Tarek Sheasha, Thomas Georgiou, Thomas Scialom, Tobias Speckbacher,
  Todor Mihaylov, Tong Xiao, Ujjwal Karn, Vedanuj Goswami, Vibhor Gupta, Vignesh Ramanathan, Viktor Kerkez, Vincent Gonguet, Virginie Do, Vish Vogeti, Vítor Albiero, Vladan Petrovic, Weiwei Chu, Wenhan Xiong, Wenyin Fu, Whitney Meers, Xavier Martinet, Xiaodong Wang, Xiaofang Wang, Xiaoqing~Ellen Tan, Xide Xia, Xinfeng Xie, Xuchao Jia, Xuewei Wang, Yaelle Goldschlag, Yashesh Gaur, Yasmine Babaei, Yi~Wen, Yiwen Song, Yuchen Zhang, Yue Li, Yuning Mao, Zacharie~Delpierre Coudert, Zheng Yan, Zhengxing Chen, Zoe Papakipos, Aaditya Singh, Aayushi Srivastava, Abha Jain, Adam Kelsey, Adam Shajnfeld, Adithya Gangidi, Adolfo Victoria, Ahuva Goldstand, Ajay Menon, Ajay Sharma, Alex Boesenberg, Alexei Baevski, Allie Feinstein, Amanda Kallet, Amit Sangani, Amos Teo, Anam Yunus, Andrei Lupu, Andres Alvarado, Andrew Caples, Andrew Gu, Andrew Ho, Andrew Poulton, Andrew Ryan, Ankit Ramchandani, Annie Dong, Annie Franco, Anuj Goyal, Aparajita Saraf, Arkabandhu Chowdhury, Ashley Gabriel, Ashwin Bharambe, Assaf Eisenman, Azadeh
  Yazdan, Beau James, Ben Maurer, Benjamin Leonhardi, Bernie Huang, Beth Loyd, Beto~De Paola, Bhargavi Paranjape, Bing Liu, Bo~Wu, Boyu Ni, Braden Hancock, Bram Wasti, Brandon Spence, Brani Stojkovic, Brian Gamido, Britt Montalvo, Carl Parker, Carly Burton, Catalina Mejia, Ce~Liu, Changhan Wang, Changkyu Kim, Chao Zhou, Chester Hu, Ching-Hsiang Chu, Chris Cai, Chris Tindal, Christoph Feichtenhofer, Cynthia Gao, Damon Civin, Dana Beaty, Daniel Kreymer, Daniel Li, David Adkins, David Xu, Davide Testuggine, Delia David, Devi Parikh, Diana Liskovich, Didem Foss, Dingkang Wang, Duc Le, Dustin Holland, Edward Dowling, Eissa Jamil, Elaine Montgomery, Eleonora Presani, Emily Hahn, Emily Wood, Eric-Tuan Le, Erik Brinkman, Esteban Arcaute, Evan Dunbar, Evan Smothers, Fei Sun, Felix Kreuk, Feng Tian, Filippos Kokkinos, Firat Ozgenel, Francesco Caggioni, Frank Kanayet, Frank Seide, Gabriela~Medina Florez, Gabriella Schwarz, Gada Badeer, Georgia Swee, Gil Halpern, Grant Herman, Grigory Sizov, Guangyi, Zhang, Guna
  Lakshminarayanan, Hakan Inan, Hamid Shojanazeri, Han Zou, Hannah Wang, Hanwen Zha, Haroun Habeeb, Harrison Rudolph, Helen Suk, Henry Aspegren, Hunter Goldman, Hongyuan Zhan, Ibrahim Damlaj, Igor Molybog, Igor Tufanov, Ilias Leontiadis, Irina-Elena Veliche, Itai Gat, Jake Weissman, James Geboski, James Kohli, Janice Lam, Japhet Asher, Jean-Baptiste Gaya, Jeff Marcus, Jeff Tang, Jennifer Chan, Jenny Zhen, Jeremy Reizenstein, Jeremy Teboul, Jessica Zhong, Jian Jin, Jingyi Yang, Joe Cummings, Jon Carvill, Jon Shepard, Jonathan McPhie, Jonathan Torres, Josh Ginsburg, Junjie Wang, Kai Wu, Kam~Hou U, Karan Saxena, Kartikay Khandelwal, Katayoun Zand, Kathy Matosich, Kaushik Veeraraghavan, Kelly Michelena, Keqian Li, Kiran Jagadeesh, Kun Huang, Kunal Chawla, Kyle Huang, Lailin Chen, Lakshya Garg, Lavender A, Leandro Silva, Lee Bell, Lei Zhang, Liangpeng Guo, Licheng Yu, Liron Moshkovich, Luca Wehrstedt, Madian Khabsa, Manav Avalani, Manish Bhatt, Martynas Mankus, Matan Hasson, Matthew Lennie, Matthias Reso, Maxim
  Groshev, Maxim Naumov, Maya Lathi, Meghan Keneally, Miao Liu, Michael~L. Seltzer, Michal Valko, Michelle Restrepo, Mihir Patel, Mik Vyatskov, Mikayel Samvelyan, Mike Clark, Mike Macey, Mike Wang, Miquel~Jubert Hermoso, Mo~Metanat, Mohammad Rastegari, Munish Bansal, Nandhini Santhanam, Natascha Parks, Natasha White, Navyata Bawa, Nayan Singhal, Nick Egebo, Nicolas Usunier, Nikhil Mehta, Nikolay~Pavlovich Laptev, Ning Dong, Norman Cheng, Oleg Chernoguz, Olivia Hart, Omkar Salpekar, Ozlem Kalinli, Parkin Kent, Parth Parekh, Paul Saab, Pavan Balaji, Pedro Rittner, Philip Bontrager, Pierre Roux, Piotr Dollar, Polina Zvyagina, Prashant Ratanchandani, Pritish Yuvraj, Qian Liang, Rachad Alao, Rachel Rodriguez, Rafi Ayub, Raghotham Murthy, Raghu Nayani, Rahul Mitra, Rangaprabhu Parthasarathy, Raymond Li, Rebekkah Hogan, Robin Battey, Rocky Wang, Russ Howes, Ruty Rinott, Sachin Mehta, Sachin Siby, Sai~Jayesh Bondu, Samyak Datta, Sara Chugh, Sara Hunt, Sargun Dhillon, Sasha Sidorov, Satadru Pan, Saurabh Mahajan,
  Saurabh Verma, Seiji Yamamoto, Sharadh Ramaswamy, Shaun Lindsay, Shaun Lindsay, Sheng Feng, Shenghao Lin, Shengxin~Cindy Zha, Shishir Patil, Shiva Shankar, Shuqiang Zhang, Shuqiang Zhang, Sinong Wang, Sneha Agarwal, Soji Sajuyigbe, Soumith Chintala, Stephanie Max, Stephen Chen, Steve Kehoe, Steve Satterfield, Sudarshan Govindaprasad, Sumit Gupta, Summer Deng, Sungmin Cho, Sunny Virk, Suraj Subramanian, Sy~Choudhury, Sydney Goldman, Tal Remez, Tamar Glaser, Tamara Best, Thilo Koehler, Thomas Robinson, Tianhe Li, Tianjun Zhang, Tim Matthews, Timothy Chou, Tzook Shaked, Varun Vontimitta, Victoria Ajayi, Victoria Montanez, Vijai Mohan, Vinay~Satish Kumar, Vishal Mangla, Vlad Ionescu, Vlad Poenaru, Vlad~Tiberiu Mihailescu, Vladimir Ivanov, Wei Li, Wenchen Wang, Wenwen Jiang, Wes Bouaziz, Will Constable, Xiaocheng Tang, Xiaojian Wu, Xiaolan Wang, Xilun Wu, Xinbo Gao, Yaniv Kleinman, Yanjun Chen, Ye~Hu, Ye~Jia, Ye~Qi, Yenda Li, Yilin Zhang, Ying Zhang, Yossi Adi, Youngjin Nam, Yu, Wang, Yu~Zhao, Yuchen Hao, Yundi
  Qian, Yunlu Li, Yuzi He, Zach Rait, Zachary DeVito, Zef Rosnbrick, Zhaoduo Wen, Zhenyu Yang, Zhiwei Zhao, and Zhiyu Ma. 2024.
\newblock \href {https://arxiv.org/abs/2407.21783} {The llama 3 herd of models}.
\newblock \emph{Preprint}, arXiv:2407.21783.

\bibitem[{Gur et~al.(2024)Gur, Furuta, Huang, Safdari, Matsuo, Eck, and Faust}]{gur2024a}
Izzeddin Gur, Hiroki Furuta, Austin~V Huang, Mustafa Safdari, Yutaka Matsuo, Douglas Eck, and Aleksandra Faust. 2024.
\newblock \href {https://openreview.net/forum?id=9JQtrumvg8} {A real-world webagent with planning, long context understanding, and program synthesis}.
\newblock In \emph{The Twelfth International Conference on Learning Representations}.

\bibitem[{Hao et~al.(2024)Hao, Chen, Zhang, and Fan}]{Yilun2024travel}
Yilun Hao, Yongchao Chen, Yang Zhang, and Chuchu Fan. 2024.
\newblock Large language models can plan your travels rigorously with formal verification tools.
\newblock \emph{arXiv preprint arXiv:2404.11891}.

\bibitem[{He-Yueya et~al.(2023)He-Yueya, Poesia, Wang, and Goodman}]{heyueya2023solvingmathwordproblems}
Joy He-Yueya, Gabriel Poesia, Rose~E. Wang, and Noah~D. Goodman. 2023.
\newblock \href {https://arxiv.org/abs/2304.09102} {Solving math word problems by combining language models with symbolic solvers}.
\newblock \emph{Preprint}, arXiv:2304.09102.

\bibitem[{Hsu et~al.(2024)Hsu, Liu, Liao, Hsu, Chen, and Shiu}]{hsu2024breeze}
Chan-Jan Hsu, Chang-Le Liu, Feng-Ting Liao, Po-Chun Hsu, Yi-Chang Chen, and Da-Shan Shiu. 2024.
\newblock Breeze-7b technical report.
\newblock \emph{arXiv preprint arXiv:2403.02712}.

\bibitem[{Huang et~al.(2024)Huang, Zhong, Lu, Zhu, Gao, Liu, Hou, Zeng, Wang, Shang, Jiang, Xu, and Liu}]{huang-etal-2024-planning-creation}
Shijue Huang, Wanjun Zhong, Jianqiao Lu, Qi~Zhu, Jiahui Gao, Weiwen Liu, Yutai Hou, Xingshan Zeng, Yasheng Wang, Lifeng Shang, Xin Jiang, Ruifeng Xu, and Qun Liu. 2024.
\newblock \href {https://doi.org/10.18653/v1/2024.findings-acl.259} {Planning, creation, usage: Benchmarking {LLM}s for comprehensive tool utilization in real-world complex scenarios}.
\newblock In \emph{Findings of the Association for Computational Linguistics: ACL 2024}, pages 4363--4400, Bangkok, Thailand. Association for Computational Linguistics.

\bibitem[{Komeili et~al.(2021)Komeili, Shuster, and Weston}]{Komeili2021InternetAugmentedDG}
Mojtaba Komeili, Kurt Shuster, and Jason Weston. 2021.
\newblock \href {https://api.semanticscholar.org/CorpusID:236034557} {Internet-augmented dialogue generation}.
\newblock In \emph{Annual Meeting of the Association for Computational Linguistics}.

\bibitem[{Lee et~al.(2024)Lee, Lin, Ho, Yu, Chen, and Shiu}]{fc-leaderboard-zhtw}
Liang-Chieh Lee, Cheng-Wei Lin, Pei-Chen Ho, Chien-Yu Yu, Yi-Chang Chen, and Da-Shan Shiu. 2024.
\newblock \href {https://github.com/mtkresearch/function-calling-leaderboard-for-zhtw} {Function calling leaderboard for zhtw}.

\bibitem[{Liu et~al.(2024{\natexlab{a}})Liu, Huang, Zeng, Hao, Yu, Li, Wang, Gan, Liu, Yu, Wang, Wang, Ning, Hou, Wang, Wu, Wang, Liu, Wang, Tang, Tu, Shang, Jiang, Tang, Lian, Liu, and Chen}]{liu2024toolacewinningpointsllm}
Weiwen Liu, Xu~Huang, Xingshan Zeng, Xinlong Hao, Shuai Yu, Dexun Li, Shuai Wang, Weinan Gan, Zhengying Liu, Yuanqing Yu, Zezhong Wang, Yuxian Wang, Wu~Ning, Yutai Hou, Bin Wang, Chuhan Wu, Xinzhi Wang, Yong Liu, Yasheng Wang, Duyu Tang, Dandan Tu, Lifeng Shang, Xin Jiang, Ruiming Tang, Defu Lian, Qun Liu, and Enhong Chen. 2024{\natexlab{a}}.
\newblock \href {https://arxiv.org/abs/2409.00920} {Toolace: Winning the points of llm function calling}.
\newblock \emph{Preprint}, arXiv:2409.00920.

\bibitem[{Liu et~al.(2024{\natexlab{b}})Liu, Hoang, Zhang, Zhu, Lan, Kokane, Tan, Yao, Liu, Feng et~al.}]{liu2024apigen}
Zuxin Liu, Thai Hoang, Jianguo Zhang, Ming Zhu, Tian Lan, Shirley Kokane, Juntao Tan, Weiran Yao, Zhiwei Liu, Yihao Feng, et~al. 2024{\natexlab{b}}.
\newblock Apigen: Automated pipeline for generating verifiable and diverse function-calling datasets.
\newblock \emph{arXiv preprint arXiv:2406.18518}.

\bibitem[{Longpre et~al.(2023)Longpre, Hou, Vu, Webson, Chung, Tay, Zhou, Le, Zoph, Wei et~al.}]{longpre2023flan}
Shayne Longpre, Le~Hou, Tu~Vu, Albert Webson, Hyung~Won Chung, Yi~Tay, Denny Zhou, Quoc~V Le, Barret Zoph, Jason Wei, et~al. 2023.
\newblock The flan collection: Designing data and methods for effective instruction tuning.
\newblock In \emph{International Conference on Machine Learning}, pages 22631--22648. PMLR.

\bibitem[{Nexusflow.ai(2023)}]{nexusraven}
Nexusflow.ai. 2023.
\newblock \href {https://nexusflow.ai/blogs/ravenv2} {Nexusraven-v2: Surpassing gpt-4 for zero-shot function calling}.

\bibitem[{Parisi et~al.(2022)Parisi, Zhao, and Fiedel}]{parisi2022talm}
Aaron Parisi, Yao Zhao, and Noah Fiedel. 2022.
\newblock \href {https://arxiv.org/abs/2205.12255} {Talm: Tool augmented language models}.
\newblock \emph{Preprint}, arXiv:2205.12255.

\bibitem[{Patil et~al.(2023)Patil, Zhang, Wang, and Gonzalez}]{patil2023gorilla}
Shishir~G. Patil, Tianjun Zhang, Xin Wang, and Joseph~E. Gonzalez. 2023.
\newblock Gorilla: Large language model connected with massive apis.
\newblock \emph{arXiv preprint arXiv:2305.15334}.

\bibitem[{Qin et~al.(2024)Qin, Liang, Ye, Zhu, Yan, Lu, Lin, Cong, Tang, Qian, Zhao, Hong, Tian, Xie, Zhou, Gerstein, dahai li, Liu, and Sun}]{qin2024toolllm}
Yujia Qin, Shihao Liang, Yining Ye, Kunlun Zhu, Lan Yan, Yaxi Lu, Yankai Lin, Xin Cong, Xiangru Tang, Bill Qian, Sihan Zhao, Lauren Hong, Runchu Tian, Ruobing Xie, Jie Zhou, Mark Gerstein, dahai li, Zhiyuan Liu, and Maosong Sun. 2024.
\newblock \href {https://openreview.net/forum?id=dHng2O0Jjr} {Tool{LLM}: Facilitating large language models to master 16000+ real-world {API}s}.
\newblock In \emph{The Twelfth International Conference on Learning Representations}.

\bibitem[{Qu et~al.(2024)Qu, Dai, Wei, Cai, Wang, Yin, Xua, and Wen}]{Changle2024tool}
Changle Qu, Sunhao Dai, Xiaochi Wei, Hengyi Cai, Shuaiqiang Wang, Dawei Yin, Jun Xua, and Ji-Rong Wen. 2024.
\newblock Tool learning with large language models: A survey.
\newblock \emph{arXiv preprint arXiv:2405.17935}.

\bibitem[{Schick et~al.(2023)Schick, Dwivedi-Yu, Dessi, Raileanu, Lomeli, Hambro, Zettlemoyer, Cancedda, and Scialom}]{schick2023toolformer}
Timo Schick, Jane Dwivedi-Yu, Roberto Dessi, Roberta Raileanu, Maria Lomeli, Eric Hambro, Luke Zettlemoyer, Nicola Cancedda, and Thomas Scialom. 2023.
\newblock \href {https://openreview.net/forum?id=Yacmpz84TH} {Toolformer: Language models can teach themselves to use tools}.
\newblock In \emph{Thirty-seventh Conference on Neural Information Processing Systems}.

\bibitem[{Shinn et~al.(2023)Shinn, Cassano, Berman, Gopinath, Narasimhan, and Yao}]{shinn2023reflexion}
Noah Shinn, Federico Cassano, Edward Berman, Ashwin Gopinath, Karthik Narasimhan, and Shunyu Yao. 2023.
\newblock \href {https://arxiv.org/abs/2303.11366} {Reflexion: Language agents with verbal reinforcement learning}.
\newblock \emph{Preprint}, arXiv:2303.11366.

\bibitem[{Tang et~al.(2023)Tang, Deng, Lin, Han, Liang, and Sun}]{tang2023toolalpaca}
Qiaoyu Tang, Ziliang Deng, Hongyu Lin, Xianpei Han, Qiao Liang, and Le~Sun. 2023.
\newblock \href {https://arxiv.org/abs/2306.05301} {Toolalpaca: Generalized tool learning for language models with 3000 simulated cases}.
\newblock \emph{Preprint}, arXiv:2306.05301.

\bibitem[{Teknium et~al.(2024)Teknium, Quesnelle, and Guang}]{teknium2024hermes3technicalreport}
Ryan Teknium, Jeffrey Quesnelle, and Chen Guang. 2024.
\newblock \href {https://arxiv.org/abs/2408.11857} {Hermes 3 technical report}.
\newblock \emph{Preprint}, arXiv:2408.11857.

\bibitem[{Theuma and Shareghi(2024)}]{theuma2024equippinglanguagemodelstool}
Adrian Theuma and Ehsan Shareghi. 2024.
\newblock \href {https://arxiv.org/abs/2401.15328} {Equipping language models with tool use capability for tabular data analysis in finance}.
\newblock \emph{Preprint}, arXiv:2401.15328.

\bibitem[{Wang et~al.(2024)Wang, Fang, Eisner, Van~Durme, and Su}]{wang-etal-2024-llms-imaginarium}
Boshi Wang, Hao Fang, Jason Eisner, Benjamin Van~Durme, and Yu~Su. 2024.
\newblock \href {https://doi.org/10.18653/v1/2024.acl-long.570} {{LLM}s in the imaginarium: Tool learning through simulated trial and error}.
\newblock In \emph{Proceedings of the 62nd Annual Meeting of the Association for Computational Linguistics (Volume 1: Long Papers)}, pages 10583--10604, Bangkok, Thailand. Association for Computational Linguistics.

\bibitem[{Wei et~al.(2022)Wei, Wang, Schuurmans, Bosma, Xia, Chi, Le, Zhou et~al.}]{wei2022chain}
Jason Wei, Xuezhi Wang, Dale Schuurmans, Maarten Bosma, Fei Xia, Ed~Chi, Quoc~V Le, Denny Zhou, et~al. 2022.
\newblock Chain-of-thought prompting elicits reasoning in large language models.
\newblock \emph{Advances in neural information processing systems}, 35:24824--24837.

\bibitem[{Xu et~al.(2023)Xu, Peng, Lei, Mukherjee, Liu, and Xu}]{xu2023rewoo}
Binfeng Xu, Zhiyuan Peng, Bowen Lei, Subhabrata Mukherjee, Yuchen Liu, and Dongkuan Xu. 2023.
\newblock \href {https://arxiv.org/abs/2305.18323} {Rewoo: Decoupling reasoning from observations for efficient augmented language models}.
\newblock \emph{Preprint}, arXiv:2305.18323.

\bibitem[{Yan et~al.(2024)Yan, Mao, Ji, Zhang, Patil, Stoica, and Gonzalez}]{berkeley-function-calling-leaderboard}
Fanjia Yan, Huanzhi Mao, Charlie Cheng-Jie Ji, Tianjun Zhang, Shishir~G. Patil, Ion Stoica, and Joseph~E. Gonzalez. 2024.
\newblock Berkeley function calling leaderboard.
\newblock \url{https://gorilla.cs.berkeley.edu/blogs/8_berkeley_function_calling_leaderboard.html}.

\bibitem[{Yang et~al.(2023{\natexlab{a}})Yang, Song, Li, Zhao, Ge, Li, and Shan}]{yang2023gpttools}
Rui Yang, Lin Song, Yanwei Li, Sijie Zhao, Yixiao Ge, Xiu Li, and Ying Shan. 2023{\natexlab{a}}.
\newblock \href {https://openreview.net/forum?id=cwjh8lqmOL} {{GPT}4tools: Teaching large language model to use tools via self-instruction}.
\newblock In \emph{Thirty-seventh Conference on Neural Information Processing Systems}.

\bibitem[{Yang et~al.(2023{\natexlab{b}})Yang, Li, Wang, Lin, Azarnasab, Ahmed, Liu, Liu, Zeng, and Wang}]{yang2023mmreact}
Zhengyuan Yang, Linjie Li, Jianfeng Wang, Kevin Lin, Ehsan Azarnasab, Faisal Ahmed, Zicheng Liu, Ce~Liu, Michael Zeng, and Lijuan Wang. 2023{\natexlab{b}}.
\newblock Mm-react: Prompting chatgpt for multimodal reasoning and action.

\bibitem[{Yao et~al.(2022)Yao, Zhao, Yu, Du, Shafran, Narasimhan, and Cao}]{yao2022react}
Shunyu Yao, Jeffrey Zhao, Dian Yu, Nan Du, Izhak Shafran, Karthik Narasimhan, and Yuan Cao. 2022.
\newblock React: Synergizing reasoning and acting in language models.
\newblock \emph{arXiv preprint arXiv:2210.03629}.

\bibitem[{Zheng et~al.(2023)Zheng, Chiang, Sheng, Zhuang, Wu, Zhuang, Lin, Li, Li, Xing et~al.}]{zheng2023judging}
Lianmin Zheng, Wei-Lin Chiang, Ying Sheng, Siyuan Zhuang, Zhanghao Wu, Yonghao Zhuang, Zi~Lin, Zhuohan Li, Dacheng Li, Eric Xing, et~al. 2023.
\newblock Judging llm-as-a-judge with mt-bench and chatbot arena.
\newblock \emph{Advances in Neural Information Processing Systems}, 36:46595--46623.

\bibitem[{Zhong et~al.(2023)Zhong, Du, Kai, Tang, Xu, Zhen, Hao, Xu, Yuan, and Yan}]{zhong2023llm4edaemergingprogresslarge}
Ruizhe Zhong, Xingbo Du, Shixiong Kai, Zhentao Tang, Siyuan Xu, Hui-Ling Zhen, Jianye Hao, Qiang Xu, Mingxuan Yuan, and Junchi Yan. 2023.
\newblock \href {https://arxiv.org/abs/2401.12224} {Llm4eda: Emerging progress in large language models for electronic design automation}.
\newblock \emph{Preprint}, arXiv:2401.12224.

\end{thebibliography}
% \bibliography{custom}

\newpage
\appendix
\onecolumn

\section{Details of pipeline for constructing reasoning descriptions}
\label{sec:appendix-reason}

The following prompt is for constructing reasoning descriptions. The provided conversation trajectory is given in \texttt{\{CONVERSATIONS\}}, the provided function descriptions are in \texttt{\{FUNCTIONS\}}, and the provided function calls are in \texttt{\{FUNC\_CALL\}}.

\begin{Verbatim}[breaklines=true, frame=single]
Your mission is to identify the reason for using the tool based on the history conversations.

## Example 1:

Given the history conversations as follows:
"""
[SYSTEM] You are a helpful assistant.
[USER] What is the weather in Taipei?
[BOT] Current temperature in Taipei: 32 Celsius
[USER] What is the weather in Palo Alto?
"""
and the available tools are as follows:
```json
[
  {
    "name": "weather_api.get_current_weather",
    "description": "Retrieves the current weather conditions for a specified location.",
    "parameters": {
      "location": {
        "type": "string",
        "description": "The name of the city or geographic location.",
        "required": true
      },
      "units": {
        "type": "string",
        "description": "The units for temperature measurement (e.g., 'Celsius', 'Fahrenheit').",
        "required": false
      }
    }
  }
]
```

Please output JSON with the key `reason` for identifying the reason 
to figure out how to use the available functions and finally expect to get the answer shown below.
```json
[
  {
    "name": "weather_api.get_current_weather",
    "arguments": {
      "location": "Palo Alto",
      "units": "Celsius"
    }
  }
]
```

## Output for Example1

```json
{
  "reason": "The user wants to know the current weather conditions in Palo Alto. The available tool 'weather_api.get_current_weather' can be used to retrieve this information by specifying the location as 'Palo Alto'."
}
```

## Example 2:

Given the history conversations as follows:
"""
[USER] Find the sum of all the multiples of 3 and 5 between 1 and 1000. Also find the product of the first five prime numbers.
"""
and the available tools are as follows:
```json
[
  {
    "name": "math_toolkit.sum_of_multiples",
    "description": "Find the sum of all multiples of specified numbers within a specified range.",
    "parameters": {
      "lower_limit": {
        "type": "integer",
        "description": "The start of the range (inclusive).",
        "required": true
      },
      "upper_limit": {
        "type": "integer",
        "description": "The end of the range (inclusive).",
        "required": true
      },
      "multiples": {
        "type": "array",
        "description": "The numbers to find multiples of.",
        "required": true
      }
    }
  },
  {
    "name": "math_toolkit.product_of_primes",
    "description": "Find the product of the first n prime numbers.",
    "parameters": {
      "count": {
        "type": "integer",
        "description": "The number of prime numbers to multiply together.",
        "required": true
      }
    }
  }
]
```

Please output JSON with the key `reason` for identifying the reason 
to figure out how to use the available functions and finally expect to get the answer shown below.
```json
[
  {
    "name": "math_toolkit.sum_of_multiples",
    "arguments": {
      "lower_limit": 1,
      "upper_limit": 1000,
      "multiples": [3, 5]
    }
  },
  {
    "name": "math_toolkit.product_of_primes",
    "arguments": {
      "count": 5
    }
  }
]
```

## Output for Example2

```json
{
  "reason": "The user wants to find the sum of all multiples of 3 and 5 between 1 and 1000, and also find the product of the first five prime numbers. The available tools 'math_toolkit.sum_of_multiples' and 'math_toolkit.product_of_primes' can be used to retrieve this information. The 'math_toolkit.sum_of_multiples' tool can be used by specifying the lower limit as 1, the upper limit as 1000, and the multiples as [3, 5]. The 'math_toolkit.product_of_primes' tool can be used by specifying the count as 5."
}
```

## Start

Given the history conversations as follows:
"""
{CONVERSATIONS}
"""
and the available tools are as follows:
```json
{FUNCTIONS}
```

Please output JSON with the key `reason` for identifying the reason 
to figure out how to use the available functions and finally expect to get the answer shown below.
```json
{FUNC_CALL}
```
\end{Verbatim}

\section{Details of pipeline for translating function-calling data}
\label{sec:appendix-translate}

The following prompt is for translating function-calling data, where the provided function-calling data in JSON format is specified in \texttt{\{DATA\}}, and the target language is indicated in \texttt{\{TARGET\_LANG\}}, e.g., "Traditional Chinese."

\begin{Verbatim}[breaklines=true, frame=single]
This JSON object outlines a conversation between a user and an assistant, including the available functions the assistant can utilize to meet the user's requests. 

In this JSON object:
- The `functions` key lists the available functions the assistant can use, including their descriptions and parameters.
- The `conversations` key outlines the conversation between the user and the assistant.
- The `tool_calls` key within the assistant's response shows the function calls the assistant makes to fulfill the user's requests, including the function name and arguments.

```json
{DATA}
```

AND NOW,
I want to translate this JSON into {TARGET_LANG}.
Note that:
- Do not translate any content in `functions`
- Translate the content in `arguments` if using {TARGET_LANG} is reasonable

Please provide your translation into JSON as same format above.
\end{Verbatim}

\end{document}